\pdfoutput=1

\documentclass[11pt]{article}

\usepackage[]{acl}

\usepackage{times}
\usepackage{latexsym}

\usepackage[T1]{fontenc}

\usepackage[utf8]{inputenc}

\usepackage{microtype}

\usepackage{graphicx}
\usepackage{amsmath}
\usepackage{multirow, epstopdf}
\usepackage{amssymb}

\title{Don’t sweat the small stuff, classify the rest: \textit{Sample Shielding} to protect text classifiers against adversarial attacks}

\author{Jonathan Rusert, Padmini Srinivasan \\
  University of Iowa\\
  \texttt{\{jonathan-rusert, padmini-srinivasan\}@uiowa.edu} \\}
\date{August 2021}

\begin{document}

\maketitle

\begin{abstract}
Deep learning (DL) is being used extensively for text classification. However, researchers have demonstrated the  vulnerability of such classifiers to adversarial attacks.  Attackers modify the text in a way which misleads the classifier while keeping the original meaning close to intact.  State-of-the-art (SOTA) attack algorithms follow the general principle of making minimal changes to the text so as to not jeopardize semantics.  
Taking advantage of this we propose a novel and intuitive defense strategy called \textit{Sample Shielding}.
It is attacker and classifier agnostic, does not require any reconfiguration of the classifier or external resources and is simple to implement. Essentially, we sample subsets of the input text, classify them and summarize these into a final decision. 
We shield three popular DL text classifiers with \textit{Sample Shielding}, test their resilience against four SOTA attackers across three datasets in a realistic threat setting.
Even when given the advantage of knowing about our shielding strategy the adversary's attack success rate is $<=10\%$ with only one exception and often $< 5\%$.  
Additionally, \textit{Sample Shielding}  maintains near original accuracy when applied to original texts.  Crucially, we show that the `make minimal changes' approach of SOTA attackers leads to critical vulnerabilities that can be defended against with an intuitive sampling strategy.\footnote{Our code and data are available at: https://github.com/JonRusert/SampleShielding}

\end{abstract}

\section{Introduction}

Text classifiers have become ubiquitous.
Unfortunately, they are subject to attacks from adversaries, typically executed using machine learning methods.
Attackers work by making small modifications to the text that mislead the classifier.
Adversarial attackers are now a growing part of the ecosystem. 

Like classifiers, attack algorithms have achieved strong success due to advances in machine learning/deep learning. Current text attackers, like TextFooler \cite{jin2020bert} and Bert-Attack \cite{li-etal-2020-bert-attack}, are able to reduce near perfect classification accuracy down to $5\%$. Additionally, these attackers achieve this while perturbing (changing) only a small amount of the original text. This helps preserve the original meaning so that humans are able to understand the original message even though classifiers are duped. 

As a counter, classifier shielding techniques are being explored.
One such approach is adversarial training where the classifier, assumed to have access to the attacker, uses it to generate perturbed texts - these are added to the classifier's training data.
While this leads to model resilience against \textit{that} attacker 
it leaves the classifier open to attacks by new attackers. 
Other defenses involve modifying classifier structure to reduce the information an attacker can glean from it \cite{goal2020DNDNet}. However, this type of reconfiguration will not be possible if a third party classifier (e.g. Google Perspective) is leveraged.
Even other approaches involve modifying the input text during classification time, but are currently limited to classifiers built from specific masked language models \cite{Zeng2021CertifiedRT} or rely on external synonym datasets \cite{Wang2021RandomizedSA}. 
We propose a shielding technique which is attacker-agnostic, does not require additional training/reconfiguration to the classifier, can shield any classifier, does not require an external data source, and can be used in a more realistic threat setting. We refer to this as \textit{Sample Shielding}.

\textit{Sample Shielding} takes advantage of current constraints in SOTA attacks. Mainly, to preserve original meaning, these make the minimal changes needed to deceive the classifier. For example, BERT-Attack \cite{li-etal-2020-bert-attack} only perturbs up to 16\% of text, and often far less (e.g. 1.1 \%) for some datasets. Thus, if we would look at the 84\% to 99\% of text that is untouched our model would be more likely to classify correctly. Hence, in \textit{Sample Shielding} we take many samples of the input text, classify these individually and combine their decisions as an ensemble to classify the text. 
Our contributions are as follows:

1. We propose a new, intuitive shielding technique called \textit{Sample Shielding} for text classifiers. 

2. We assess \textit{Sample Shielding} under a realistic threat model where the attacker cannot query a website's classifier hundreds of times since that pattern is easily detectable by the website. We run experiments under two conditions, when the attacker has knowledge of \textit{Sample Shielding} and when it does not. In both cases the attacker uses a local copy of the websites' classifier.
This is an optimistic assumption favouring the attacker and thus provides a lower bound to our results.

3. We test against 4 SOTA text attack algorithms, 3 text datasets and 3 classifiers. When the attacker does not have knowledge of \textit{Sample Shielding},  our defense reduces attack success rate from near total decimation 90 - 100\% down to 13 - 36\%, while still maintaining accuracy on original texts. When the attacker has knowledge of \textit{Sample Shielding}, our defense performs even better, reducing attacks down to 1 - 10\% success rate. This is partially due to \textit{Sample Shielding's} random nature providing unreliable feedback to attackers.

 
%

Our success with \textit{Sample Shielding}
is good news for classifiers -- and it raises the bar significantly for the next generation attackers.
We share code and our perturbed text collections for future research.


\section{Methodology}

\subsection{Threat model}

\begin{figure}
    \centering
    \includegraphics[width=0.8\columnwidth]{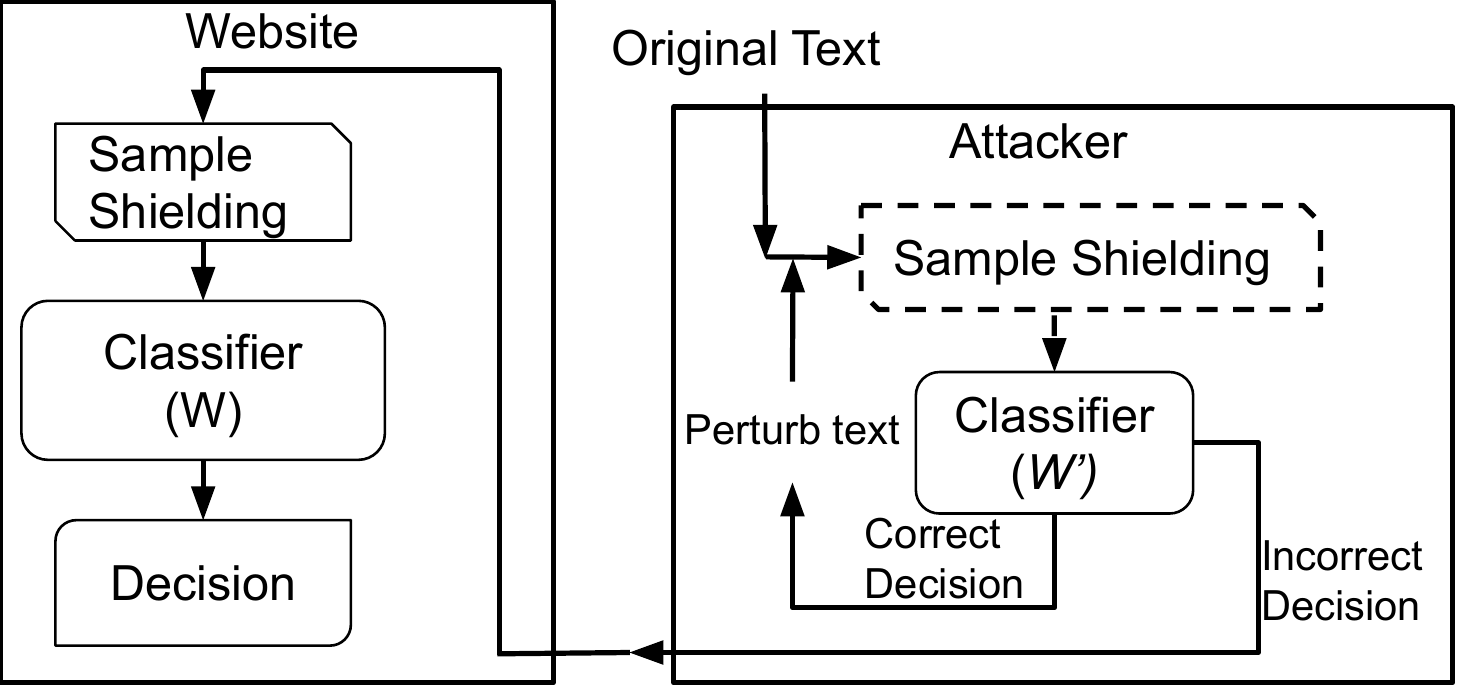}
    \caption{Threat model - Attacker modifies text with feedback from its local classifier $W'$. Dashed box included in path when attacker knows about \emph{Sample Shielding} employed by website. When box excluded  knowledge of \emph{Sample Shielding} is unavailable.}
    \label{fig:threatmodel}
\end{figure}

The typical attack strategy perturbing texts with word synonyms or character substitutions assumes to have query access to the target web site's classifier ($W$) \cite{yoo2021improving,li2021contextualized,ren-etal-2019-generating,jin2020bert,li-etal-2020-bert-attack,garg2020bae,jia2019certified,Li2019Textbugger}.  
The text is modified by querying $W$ hundreds or thousands of times, each time with a text version differing only slightly from the previous - even by just a single word  \cite{li-etal-2020-bert-attack,jin2020bert}.
Such a querying pattern can be easily identified as adversarial by the website and countered.
Thus, practically the only way in which such an attack can take place is when the attacker owns a local classifier $W'$ which is either an exact copy of $W$ or a close enough approximation.
We adopt this more realistic threat model, shown in Figure \ref{fig:threatmodel}. 

In our threat model the attacker uses feedback from its local $W'$ to generate a final perturbed version that defeats $W'$ or is close enough to do so.  The attacker submits only this final version to the website, expecting $W$ to make the same error.
%
However, the website defends $W$ using \textit{Sample Shielding}: sample based pre-processing on the input text, prior to applying $W$. The attacker may or may not be aware of this fact. 
Keeping $W = W'$ which is consistent with other defenses, we evaluate our defense under two conditions:

1)
The attacker does not know that the website employs \textit{Sample Shielding} pre-processing when classifying text using  $W$.

2) The \textit{Sample Shielding} step is leaked and the attacker incorporates it locally when using $W'$ to generate the final perturbed text.





We present results from experiments  exploring both of these attack conditions.


\subsection{\textit{Sample  Shielding} approach}

\begin{figure*}
    \centering
    \includegraphics[width=0.75\linewidth]{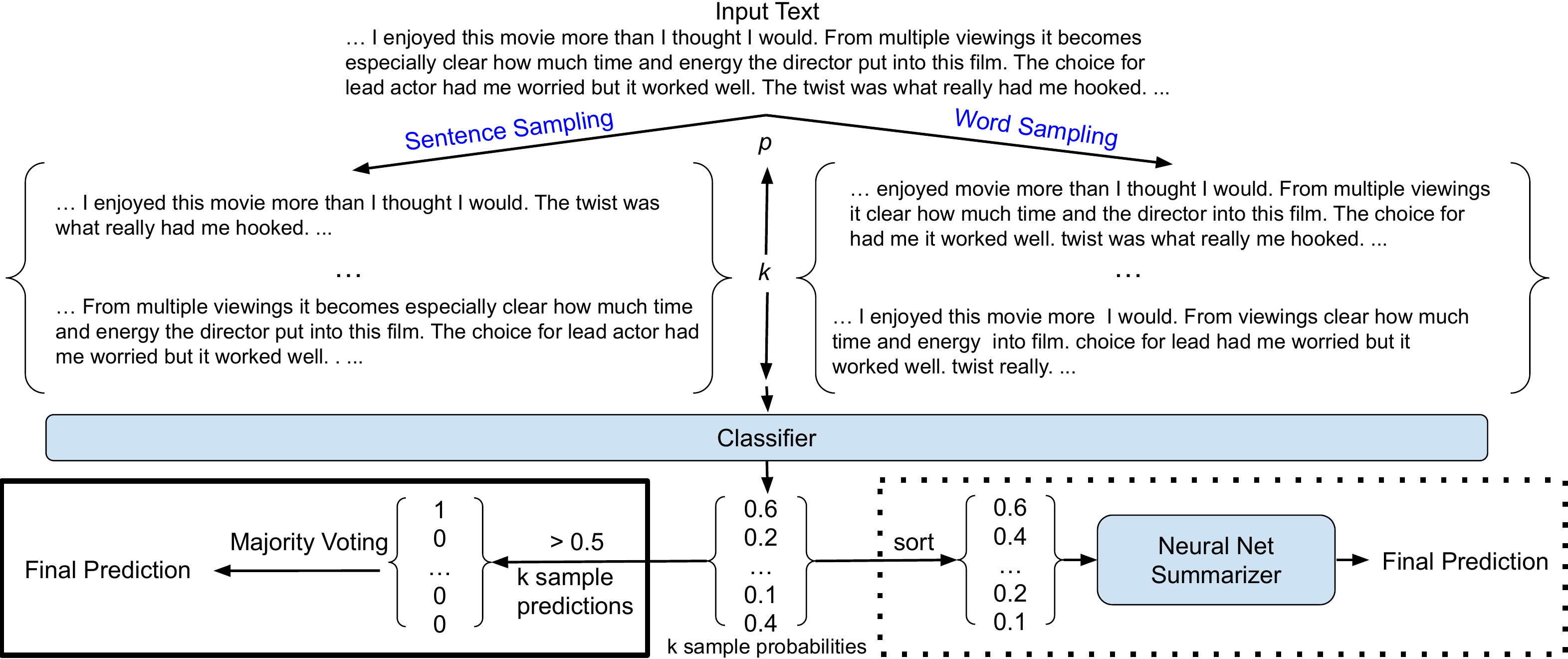}
    \caption{Proposed shielding method. Sentences or words are sampled $k$ times at a rate of $p$ percent (of the input text), the $k$ samples are classified. The probabilities are used in a  majority vote for the final prediction (solid box), or are sorted and given to a Neural Net Summarizer (NN or NN-BB) to made the final prediction (dotted box).}
    \label{fig:sampleshielding}
\end{figure*}

\noindent 
\textbf{ Intuition.} Current adversarial attackers have two goals: fool the classifier and maintain the original meaning. %
Since they make minimal changes, the extent of perturbation is in fact one of the reported statistics.
For example, \cite{li-etal-2020-bert-attack} note that their  10\% perturbation rate is far less than in previous attacks. \cite{Li2019Textbugger} also focus on minimal changes (4\%) needed in support of their attack success rate. Our defense approach capitalizes on this drive to make minimal changes. Specifically, in \textit{Sample Shielding}, we take $k$ samples each composed of $p\%$ of the text. We choose a $p$ which minimizes the chance of a sample including attacked (modified) words, while maximizing the content available for the classifier to make a correct classification. We choose a $k$ which is large enough to cover key  information but small enough to reduce redundancy. We classify each sample and combine their decisions for the final classification. We explore two sampling  and three decision combining methods.

\subsubsection{Sampling methods}
\label{sec:sampling}

\textbf{Random Sampling.}
We randomly sample $p$ portions of the text.
We explore both sentences and words as sampled units.
A visualization of random sampling is in Figure \ref{fig:sampleshielding}.

\noindent
\textbf{Shifting Sampling.}
We sample the text using a moving window of length $p \times length\_of\_text$.
The first starts at the beginning of the text.
The next window starts right after the previous window ends. If there is insufficient text for the last window, then it wraps back to include the beginning text. 

\subsubsection{Decision strategy}

\noindent
\textbf{Majority voting.}
This is a simple majority vote across the k samples (Figure \ref{fig:sampleshielding}).

\noindent
\noindent \textbf{Classifier trained on sample scores from original texts (NN).}
We train a neural network summarizer to make a final class prediction based on the $k$ sample probabilities.
Since sample ID does not carry any information,
the input to the neural network is a sorted list of sample probabilities.
The intent is to see if the neural network picks up on latent patterns in the probabilities that are not captured by majority voting (see Figure \ref{fig:sampleshielding}). It should be emphasized that the neural network summarizer is trained only on probabilities generated from original texts and does not consider probabilities from attacker modified texts. We use a simple feed forward neural net composed of 2 linear layers (size 500 and 300) as classification summarizer. 

\noindent
\textbf{Classifier trained on sample scores from original and attacked texts (NN-BB).} 
This is similar to the previous strategy except that the training data includes scores from original texts and texts that have been modified by the attacker. Because this assumes more knowledge of the attacker we expect NN-BB to perform better than NN.
The ground truth label for these modified texts is the original correct class label.

\section{Experimental Setup}

\subsection{Datasets}

We examine three standard datasets in our experiments.
Two have binary class labels (Yelp, IMDB) and the third has multi class labels (AG News).
These have been used in adversarial generation and defense research \cite{Zeng2021CertifiedRT, li-etal-2020-bert-attack}. 
All datasets can be found via huggingface\footnote{huggingface.co/datasets}.

1. IMDB - Movie review dataset for binary sentiment classification. 25k examples are provided for training and testing respectively.

2. Yelp - Yelp dataset for binary sentiment classification on reviews of businesses extracted from the Yelp Dataset Challenge\footnote{www.yelp.com/dataset/challenge/winners}. 560k examples are provided for training and 38k for testing.

3. AG News - News articles from over 2000 news sources annotated by type of news: Sports, World, Business, and Science/Tech. 120k training and 7k test sets are provided.

Following previous research, \cite{li-etal-2020-bert-attack, jin2020bert} we use all training data, and evaluate our method on random 1k samples of each dataset for the case where the local classifier does not employ \textit{Sample Shielding}. Due to the high amount of queries used by the adversaries, we test on a subset of 100 samples for the case where the attacker's local classifier employs \textit{Sample Shielding}.\footnote{We share the original and perturbed texts for replicability. We note that replicability of previous defenses are limited because the identity of their randomly sampled test instances  are not provided. }

\subsection{Adversarial models}

We test our text classifier shielding strategy against 4 state-of-the-art (SOTA) text classifier attack algorithms. These algorithms have shown excellent performance in causing misclassifications while still producing readable texts. We defend against 3 word based attacks: TextFooler \cite{jin2020bert}, Bert-Attack \cite{li-etal-2020-bert-attack}, PWWS \cite{ren-etal-2019-generating}. TextFooler leverages word embeddings for word replacements, Bert-Attack leverages BERT itself by masking words and using BERT suggestions, PWWS selects and weights word replacements from WordNet. All three use some form of greedy selection for determining which words to replace. We also defend against a character based attack algorithm, TextBugger \cite{Li2019Textbugger}.

\subsection{ Victim classifier models}

We test our  shielding approach against 3 standard classifiers\footnote{We calibrated classifier accuracies against previous research \cite{li-etal-2020-bert-attack, jin2020bert}} used in previous research, e.g. \cite{li2021contextualized, jin2020bert, li-etal-2020-bert-attack}:

1. CNN - A word based CNN \cite{kim-2014-convolutional}, with three window sizes (3,4,5), 100 filters per window with dropout of 0.3 and Glove embeddings.

2. LSTM - A word based bidirectional LSTM with 150 hidden units. As with the CNN a dropout of 0.3 is used and Glove embeddings are leveraged.

3. BERT - The 12 layer BERT base model which has been fine-tuned on the corresponding dataset. These are provided by textattack via huggingface\footnote{huggingface.co/textattack}.

\subsection{Experimental design}

We run experiments on the combination of the three victim classification models, three datasets, and four attack algorithms. These combinations are run on both threat model conditions (attacker is aware/ not aware of SampleShielding). This leads to 72 shielding experiments. 
For all attacks, we leverage TextAttack framework\footnote{textattack.readthedocs.io/en/latest/index.html} which provides classification algorithms and adversarial text generation algorithms implemented as specified in respective papers \cite{morris2020textattack}. 
 In all experiments where the attacker does not use \textit{Sample Shielding} we set $k=100$ and $p=0.3$.  While better performance was achieved with other values in preliminary experiments,  we chose to go with a single combination of $p$ and $k$ for simplicity.
In experiments where the attacker uses \textit{Sample Shielding} pre-processing we reduce $k$ to $30$ for efficiency.
Except where otherwise noted, majority voting is used to generate results.
 Additionally, shifting sampling (Section \ref{sec:sampling}) shielding typically achieved 10-20 points lower accuracy compared to random, thus we do not include it in the results.

\subsection{Evaluation measures}

We examine accuracy and Attack Success Rate:
\begin{equation}
    accuracy = \frac{\# examples\_classified\_correctly}{\# total\_examples}
\end{equation}

\noindent

\begin{equation}
    ASR = \frac{Original_{Acc.}- Attacked_{Acc.}}{ Original_{Acc.}}
\end{equation}


\section{Results}

We first present results for the condition where the attacker is not aware of \textit{Sample Shielding} based pre-processing and then the results for when the attacker also employs \textit{Sample Shielding}.

\begin{figure}
    \centering
    \includegraphics[width=0.7\columnwidth]{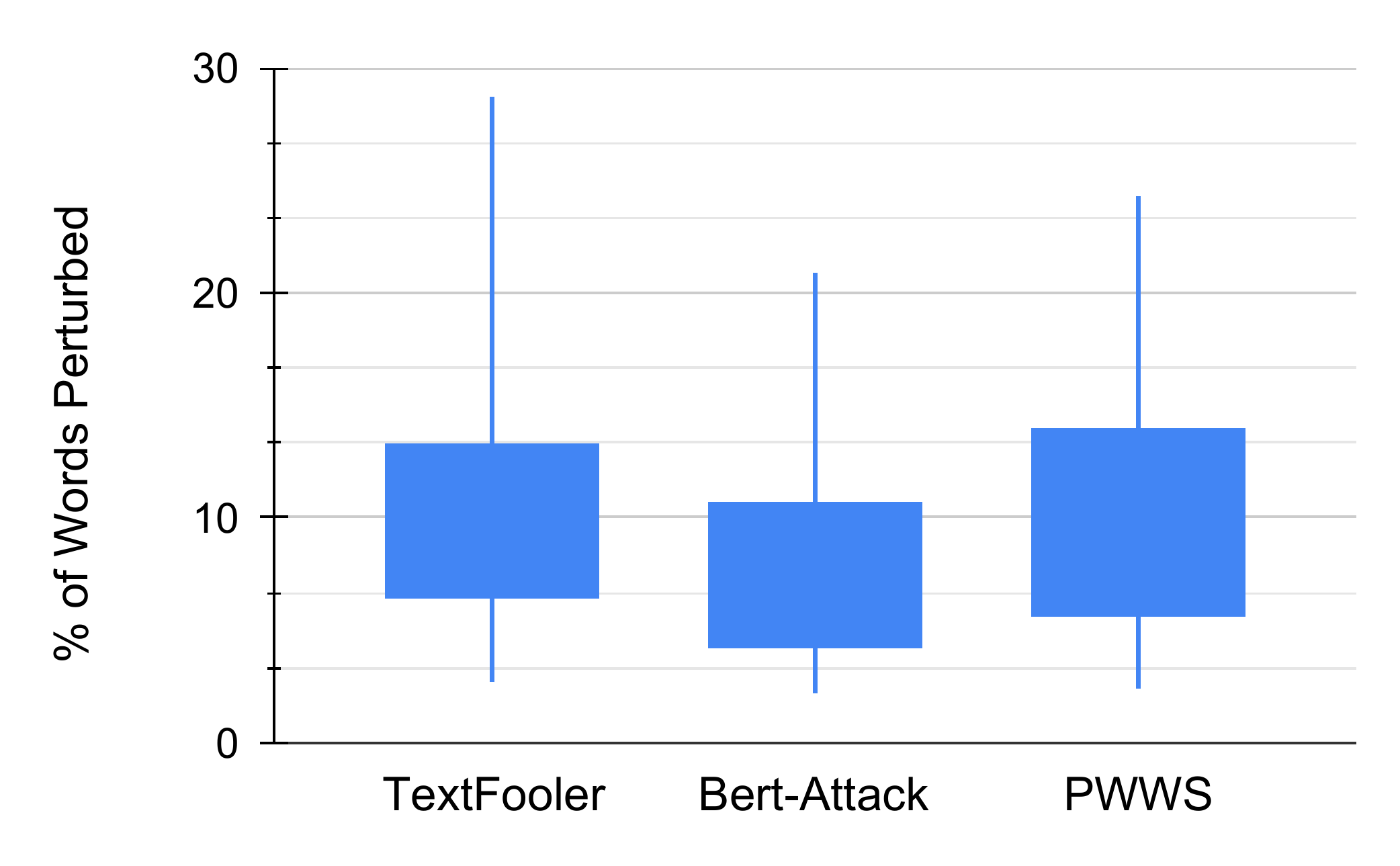}
    \caption{Average \% of perturbed words for each attack.  Percentages estimated by comparing words in original and perturbed texts. Since TextBugger adds  whitespace in words skewing its percentage it is excluded.}
    \label{fig:perturbedPercents}
\end{figure}

\subsection{Condition 1: Attacker does not know about \textit{Sample Shielding}}

Results are in Table \ref{tab:mainResults}. BERT is the strongest classifier achieving 91 - 100\% accuracy on the original datasets. 
Attacks are highly successful against unshielded texts. TextFooler and Bert-Attack are the most successful, dropping accuracies to 0-5\% generally.  Attacks were able to achieve strong drops with minimal amount of text perturbed (about 10\%). 
Figure \ref{fig:perturbedPercents} shows that the average percent of words perturbed across datasets for each attack are about equal in the mid regions of the plots.
For AG News, attacks are less successful against BERT; accuracy drops to 19\% in the strongest attack (TextFooler), and only to 49\% in the weakest (TextBugger). In general, TextBugger, the character-based attacker, is the least effective attacker. 

\begin{table*}[]
    \centering
    \footnotesize
    \begin{tabular}{c||c||c|c||c|c||c|c||c|c||c|c}
        & & Sample & Orig. & \multicolumn{2}{|c|}{TextFooler} & \multicolumn{2}{|c|}{Bert-Attack} & \multicolumn{2}{|c|}{TextBugger} & \multicolumn{2}{|c}{PWWS} \\
         & Classifier & Shielding & Acc. & Acc. & ASR  & Acc. & ASR & Acc. & ASR & Acc.  & ASR \\\hline
         \parbox[t]{2mm}{\multirow{9}{*}{\rotatebox[origin=c]{90}{IMDB}}}  & \multirow{3}{*}{LSTM} & No Shielding  & 88.3 & 0 & 100 & 0 & 100  & 0.3 & 100 & 0.1 & 100\\
         & & Shielding-Sentence & 85.1  & 61.4 & 30 & 62.0 & 30 & 60.3 & 32& 56.2 & 36 
         \\
         & & Shielding-Word & 85.1  & \textbf{66.0} & \textbf{25} & \textbf{67.0} & \textbf{24} & \textbf{66.0} & \textbf{25} & \textbf{65.7} & \textbf{26}
         \\\cline{2-12}
         
         & \multirow{3}{*}{CNN} & No Shielding  & 86.2 & 0.1 & 100 & 0 & 100 & 0.3 & 100 & 0 & 100 \\
         & & Shielding-Sentence & 84.5 & 55.3 & 36 & 55.2 & 36 & 53.6 & 38 & 48.9 & 43 \\
         & & Shielding-Word & 84.7 & \textbf{69.8} & \textbf{19} & \textbf{66.7} & \textbf{23} & \textbf{71.6} & \textbf{17} & \textbf{67.8} & \textbf{21}\\\cline{2-12}
         
         & \multirow{3}{*}{BERT} & No Shielding  & 91.3 & 1 & 99 & 3.7 & 96 & 9.2 & 90 & 0.7 & 99 \\
         & & Shielding-Sentence & 91.5 & \textbf{78.1} & \textbf{14} &\textbf{79.2} & \textbf{13} & \textbf{80.1} & \textbf{12} & \textbf{78.0} & \textbf{15} \\
         & & Shielding-Word & 86.8 & 74.4 & 19 & 71.5 & 22 & 78.8 & 14 & 63.4 & 31 \\\hline
         
         \parbox[t]{2mm}{\multirow{9}{*}{\rotatebox[origin=c]{90}{Yelp}}}  & \multirow{3}{*}{LSTM} & No Shielding  & 92.5 & 0.3 & 100 &  0.7 & 99 & 5 & 95 & 1.5 & 98 \\
         & & Shielding-Sentence & 90.0 & 62.3 & 33 & 61.1 & 34 & 60.5 & 35 & 58 & 37\\
         & & Shielding-Word & 87.8 & \textbf{65.5} & \textbf{29} & \textbf{66.7} & \textbf{28} & \textbf{68.5} & \textbf{26} & \textbf{61.9} & \textbf{33} \\\cline{2-12}
         
         & \multirow{3}{*}{CNN} & No Shielding  & 94.1 & 0.8 & 99 & 0.4 & 100 & 5.5 & 94 & 2.4 & 97\\
         & & Shielding-Sentence & 91.7 & 58.5 & 38 & 54.1 & 43 & 57.1 & 39 & 50 & 47\\
         & & Shielding-Word & 88.1 & \textbf{64.9} & \textbf{31} & \textbf{62.2} & \textbf{34} & \textbf{70.4} & \textbf{25} & \textbf{60.2} & \textbf{36} \\\cline{2-12}
         
         & \multirow{3}{*}{BERT} & No Shielding  & 100 & 5.9 & 94 & 8.3 & 92 & 15.5 & 85 & 4.9 & 95 \\
         & & Shielding-Sentence & 98.6 & \textbf{74.8} & \textbf{25} & 72.6 & 27 & \textbf{79.3} & \textbf{21} & 68.5 & 32 \\
         & & Shielding-Word & 93.5 & 69.9 & 30 & \textbf{75.1} & \textbf{25} & 78.7 & \textbf{21} & \textbf{71.1} & \textbf{29} \\\hline

         \parbox[t]{2mm}{\multirow{9}{*}{\rotatebox[origin=c]{90}{AG News}}}  & \multirow{3}{*}{LSTM} & No Shielding  & 91.6 & 1.2 & 99 & 0.9 & 99 & 16.7 & 82 & 15.6 & 83 \\
         & & Shielding-Sentence & 88.8 & 16.5 & 82 & 12.9 & 86 & 27.3 & 70 & 25.2 & 72 \\
         & & Shielding-Word & 85.1 & \textbf{60.8} & \textbf{34} & \textbf{60.9} & \textbf{34} & \textbf{60.5} & \textbf{34} & \textbf{63.7} & \textbf{30}  \\\cline{2-12}
         
         & \multirow{3}{*}{CNN} & No Shielding  & 91.5 & 0.4 & 100 & 0.3 & 100 & 5.2 & 94 & 6.3 & 93 \\
         & & Shielding-Sentence & 89.4 & 13.2 & 86 & 13.0 & 86 & 17.2 & 81 & 15.7 & 83 \\
         & & Shielding-Word & 87.8 & \textbf{77.3} & \textbf{16} & \textbf{67.7} & \textbf{26} & \textbf{74.2} & \textbf{19} & \textbf{80} & \textbf{13} 
         \\\cline{2-12}
         
         & \multirow{3}{*}{BERT} & No Shielding  & 99.6 & 18.7 & 81 & 22.5 & 77 & 49.4 & 50 & 38.5 & 61 \\
         & & Shielding-Sentence & 96.4 & 29.6 & 70 & 37.9 & 62 & 54.2 & 46 & 47.1 & 53 \\
         & & Shielding-Word & 94.5 & \textbf{75.5} & \textbf{24} & \textbf{72.0} & \textbf{28} & \textbf{78.1} & \textbf{22} & \textbf{70.5} & \textbf{29} \\\hline
         
    \end{tabular}
    \caption{Results where attacker does not know about \textit{Sample Shielding}. Shielding settings: $k=100$, $p=0.3$,  majority voting. Acc: accuracy, ASR: success rate of attack (\%), Orig. Acc.: accuracy on original texts.}
    \label{tab:mainResults}
\end{table*}

\begin{table*}[]
    \centering
    \footnotesize
    \begin{tabular}{c||c||c|c||c|c||c|c||c|c||c|c}
        & & Sampling & Orig. & \multicolumn{2}{|c|}{TextFooler} & \multicolumn{2}{|c|}{Bert-Attack} & \multicolumn{2}{|c|}{TextBugger} & \multicolumn{2}{|c}{PWWS} \\
         & Classifier & Strategy & Acc. & Acc. & SR  & Acc. & SR & Acc. & SR & Acc.  & SR \\\hline
         \parbox[t]{2mm}{\multirow{8}{*}{\rotatebox[origin=c]{90}{IMDB}}}  &
         \multirow{4}{*}{LSTM} & No Shielding&  88.3 & 0 & 100 & 0 & 100  & 0.3 & 100 & 0.1 & 100\\
         & & Maj. Vot. & 85.1  & \textbf{66.0} & \textbf{25} & 67.0 & 24 & 66.0 & \textbf{25} & 65.7 & 26
         \\
        & & NN & 85.3 & 62.5 & 29 & 62.1 & 30  & 65.4 & 26 & 62.4 & 29\\
        & & NN-BB & 85.3 & 65.2 & 26 & \textbf{68.2} & \textbf{23} & \textbf{66.5} & \textbf{25} & \textbf{67.3} & \textbf{24} \\\cline{2-12}

         & \multirow{4}{*}{CNN} & No Shielding &  86.2 & 0.1 & 100 & 0 & 100 & 0.3 & 100 & 0 & 100 \\
         & & Maj. Vot. & 84.7 &  \textbf{69.8} &  \textbf{19} & 66.7 & 23 & 71.6 & 17 & 67.8 & 21\\
         & & NN & 84.8 & 61.7 & 28 & 59.6 & 31 & 66.7 & 23 &  60.0 & 30 \\
         & & NN-BB & 84.8 & 69.3 & 20 &  \textbf{67.9} &  \textbf{21} & \textbf{72.3} & \textbf{16} & \textbf{69.6} & \textbf{19} \\\hline
         

         \parbox[t]{2mm}{\multirow{8}{*}{\rotatebox[origin=c]{90}{Yelp}}}  & \multirow{4}{*}{LSTM} & No Shielding &  92.5 & 0.3 & 100 &  0.7 & 99 & 5 & 95 & 1.5 & 98 \\
         & & Maj. Vot. & 87.8 & 65.5 & 29 & 66.7 & 28 & 68.5 & 26 & 61.9 & 33 \\
         & & NN & 89.0 & 68.7 & 26 & 68.1 & 26 &  \textbf{73.5} &  \textbf{21} & 63.6 & 31 \\
        & & NN-BB & 89 & \textbf{69.7} & \textbf{25} & \textbf{70.0} & \textbf{24} &   \textbf{73.5} &   \textbf{21} & \textbf{64.9} & \textbf{30} \\\cline{2-12}
           
         & \multirow{4}{*}{CNN} & No Shielding & 94.1 & 0.8 & 99 & 0.4 & 100 & 5.5 & 94 & 2.4 & 97\\
         & & Maj. Vot. & 88.1 & 64.9 & 31 & 62.2 & 34 & 70.4 & 25 & 60.2 & 36 \\
         & & NN & 89.9 & 63.2 & 33 & 57.6 & 39 & 69.9 & 26 & 57.4 & 39\\
         & & NN-BB & 89.9 & \textbf{72.2} & \textbf{23} & \textbf{69.7} & \textbf{26} & \textbf{72.9} & \textbf{23} &   \textbf{67.6} &   \textbf{28} \\\hline
         

    \end{tabular}
    \caption{Comparing vote summarizers. Settings: $k=100$, $p=0.3$, word sampling. Maj. Vot: majority voting, NN: neural network trained on original texts, NN-BB: neural network trained on original + perturbed texts.}
    \label{tab:secondaryResults}
\end{table*}

\noindent \textbf{\textit{Sample Shielding} greatly reduces effectiveness of attacks while maintaining accuracy on original texts.} 
The shielded classifier $W$ maintains accuracy on original texts to within 
7\% of the original accuracy. 
 Crucially, for attacked texts we see accuracy improve to between 60 and 80\% (from post attack range of 0-5\% generally). For example, TextFooler causes BERT's accuracy to drop from 91\% to 1\% for IMDB, however, \textit{Sample Shielding} returns accuracy to 78\%. In other words, the effectiveness of the attack is reduced from 99\% effective to 14\% effective. Additionally, accuracy on the original texts is maintained (91.3 to 91.5). This pattern is seen in the other attack classifier models and dataset combinations as well. For Yelp, LSTM drops from 92.5 to 0.7 when attacked by BERT-Attack, however, Word sampling brings it back up to 66.7, while achieving an original accuracy of 87.8. Overall, accuracy after shielding ranges from 60 to 80\% (avg: 70), which corresponds to a 13 - 36 (avg: 25) attack success rate.

\noindent \textbf{\textit{Sample Shielding} effective against both word based and character based attacks.} The results show effectiveness regardless of type of attack (word or character based). For example, all 4 attacks bring the original accuracy of LSTM from 88.3 down to $\thicksim $0 for IMDB. However, word sampling brings the accuracy back up to $\thicksim$66. This is a great reduction in attack effectiveness. Again, similar trends are seen for the other classifiers, CNN is reduced from 94.1 to $\leq$5.5 for Yelp, but word sampling brings it back up to 60 - 70\%.

\noindent \textbf{Word sampling outperforms sentence sampling for LSTM, CNN, sentence sampling better for BERT.} For example, for CNN on IMDB, word sampling increases accuracy more than 15 points over sentence sampling (69.8 vs 53.3). Similar trends hold for LSTM. However, the opposite is seen for BERT classifiers. For BERT on IMDB, we see an average of 6.5 higher points for sentence sampling over word sampling. These results are not surprising as LSTM and CNN leverage word embeddings for classification, while BERT leverages the context of the entire sentence.

\noindent \textbf{Word sampling is more appropriate for short texts.} With AG news, we see a large drop in effectiveness of sentence sampling. The average length of AG News is 43 words compared to 157 and 215 words of Yelp and IMDB respectively \cite{li-etal-2020-bert-attack}. This shorter length makes it more difficult to sample enough sentences. For Textfooler - CNN, sentence sampling is only able to increase accuracy from the attacked value of 0.4 to 13.2. However, word sampling is much more effective, increasing accuracy to 77.3. Text length may be crucial when choosing between the two strategies for a dataset.

\noindent \textbf{Neural Network summarizer shows some improvements over majority voting.} Comparisons of majority voting and the two neural net-based decision strategies are in Table \ref{tab:secondaryResults}. We experimented on the two binary datasets\footnote{AG News was not included due to the complexity of translating multiple probabilities to a single input.}. Replacing majority voting with a simple neural net (NN) gave somewhat disappointing results - accuracies stay  the same or decrease slightly in all cases except for LSTM on the Yelp dataset (increases).  However, when the neural nets are trained on perturbed texts (NN-BB), we see increases. For example, CNN vs TextFooler on Yelp, the neural net increases accuracy from 64.9 to 72.2, reducing attack success rate from 31 to 23. Possibly a more sophisticated neural net, such as a sequence aware LSTM, might better exploit patterns in the sorted probabilities.

\subsection{Condition 2: Attacker knows about \textit{Sample Shielding}}

Results are in Table \ref{tab:adversarialResults}. As in the previous condition, classifiers perform well on original texts (Table \ref{tab:mainResults}) with BERT often achieving the highest accuracies. In this setting, every query by an attacker requires $k$ samples to be processed, which greatly increases attack time. Thus, we reduce $k$ to 30 for these experiments.  


\begin{table*}[]
    \centering
    \footnotesize
    \begin{tabular}{c||c||c|c||c|c||c|c||c|c||c|c}
        & & Sample & Orig. & \multicolumn{2}{|c|}{TextFooler} & \multicolumn{2}{|c|}{Bert-Attack} & \multicolumn{2}{|c|}{TextBugger} & \multicolumn{2}{|c}{PWWS} \\
         & Classifier & Strategy & Acc. & Acc. & ASR  & Acc. & ASR & Acc. & ASR & Acc.  & ASR \\\hline
         \parbox[t]{2mm}{\multirow{6}{*}{\rotatebox[origin=c]{90}{IMDB}}}  &  \multirow{2}{*}{LSTM} & No Shielding  & 91	& 0 & 100 & 0 & 100 &	0 & 100 & 0 & 100 \\
          & & Shielding-Word & 94 & 89 & 5	& 87 & 7	& 89 & 5 & 89 & 5 \\\cline{2-12}
         
         & \multirow{2}{*}{CNN} & No Shielding  & 86 & 0 & 100 & 0 & 100 & 0 & 100 & 0 & 100 \\
        & & Shielding-Word & 89	& 88 & 1	& 88 & 1 & 89 & 0 & 86 & 3\\\cline{2-12}
         
         & \multirow{2}{*}{BERT} & No Shielding  & 90 & 1 & 99	& 4	& 96 & 6 & 93 & 2 & 98\\
        & & Shielding-Word & 85 & 80	& 6 & 80 & 6 & 84 & 1 & 82 & 4\\\hline 
         
          \parbox[t]{2mm}{\multirow{6}{*}{\rotatebox[origin=c]{90}{Yelp}}}  & \multirow{2}{*}{LSTM} & No Shielding  & 95 & 0 & 100 & 0 & 100 & 6 & 94 & 0 & 100\\
         & & Shielding-Word & 87 & 81& 7 & 79 & 9 & 78 & 10 & 74 & 15\\\cline{2-12}
         
         & \multirow{2}{*}{CNN} & No Shielding  & 96 & 0 & 100 & 0 & 100 & 5 & 95 & 3 & 97\\
       & & Shielding-Word & 88	& 85 & 3 & 81 & 8 & 81 & 8 & 83 & 6 \\\cline{2-12}
         
         & \multirow{2}{*}{BERT} & No Shielding  & 100	& 3	& 97 & 10 & 90 & 13 & 87 &7 & 93\\
         & & Shielding-Word & 92 & 90	& 2 & 88 & 4	& 91 & 1 & 85 & 8 \\\hline

          \parbox[t]{2mm}{\multirow{6}{*}{\rotatebox[origin=c]{90}{AG News}}} 
         & \multirow{2}{*}{LSTM} & No Shielding  & 93 &	1 & 99 & 0 & 100 & 16 & 83 & 13 & 86 \\
          & & Shielding-Word & 87 & 78 & 10	& 84 & 3 & 78 & 10 & 84 & 3\\\cline{2-12}
         
         & \multirow{2}{*}{CNN} & No Shielding  & 92 &	1 & 99 & 0 & 100 & 7 & 92 & 3 & 97\\
        & & Shielding-Word & 87	& 81 & 7	& 87 & 0 & 84 & 3 & 83 & 5
         \\\cline{2-12} 
         
         & \multirow{2}{*}{BERT} & No Shielding  & 99 & 20 & 78 & 11 & 89 & 60 & 39 & 15 & 85 \\
       & & Shielding-Word & 88	& 81 & 8 & 82 & 7 &83 & 6 & 85 & 3 \\\hline 
         
    \end{tabular}
    \caption{Results where attacker knows about \textit{Sample Shielding}. Shielding settings: $k=30$, $p=0.3$, majority voting. Acc: accuracy, ASR: success rate of attack (\%), Orig. Acc: accuracy on original texts. }
    \label{tab:adversarialResults}
\end{table*}

\noindent \textbf{\textit{Sample Shielding} repels attacks even when attacker uses \textit{Sample Shielding}.} We see that shielding is extremely successful in almost completely removing the negative effects of the attacks.
For example, on the IMDB - TextFooler combination, attack success rate drops from 100 to 5 for LSTM, 100 to 1 for CNN, and 99 to 6 against BERT. 
The largest protection provided by \textit{Sample Shielding} (100\%) is for TextBugger vs CNN in IMDB. The smallest is for 85\% (PWWS vs LSTM). On average the protection is 88.8\%.
The recovered accuracies are only 13 to 0 percent away from the originals.
These results show the power of \textit{Sample Shielding} as even with knowledge of both the classifier and \textit{Sample Shielding}, attacks struggle to perturb the text in a manner that causes $W$ to fail.
Furthermore, the attacks do worse with feedback from \textit{Sample Shielding}. This shows the misleading nature of feedback from \textit{Sample Shielding}, and unreliability when guiding attacks. 


\section{Additional Analysis}

\subsection{Parameter search}

Increasing $p$ raises the risk of samples containing increased amounts of perturbed text. Decreasing $k$ raises the risk of not covering enough of the unperturbed portions of the original text. While our settings of $p = 0.3$ and  $k = 100$ for our main results are reasonable values (Table \ref{tab:mainResults}, Table \ref{tab:secondaryResults}) they are not necessarily optimal. 

\begin{figure}
    \centering
    \includegraphics[width=0.8\columnwidth]{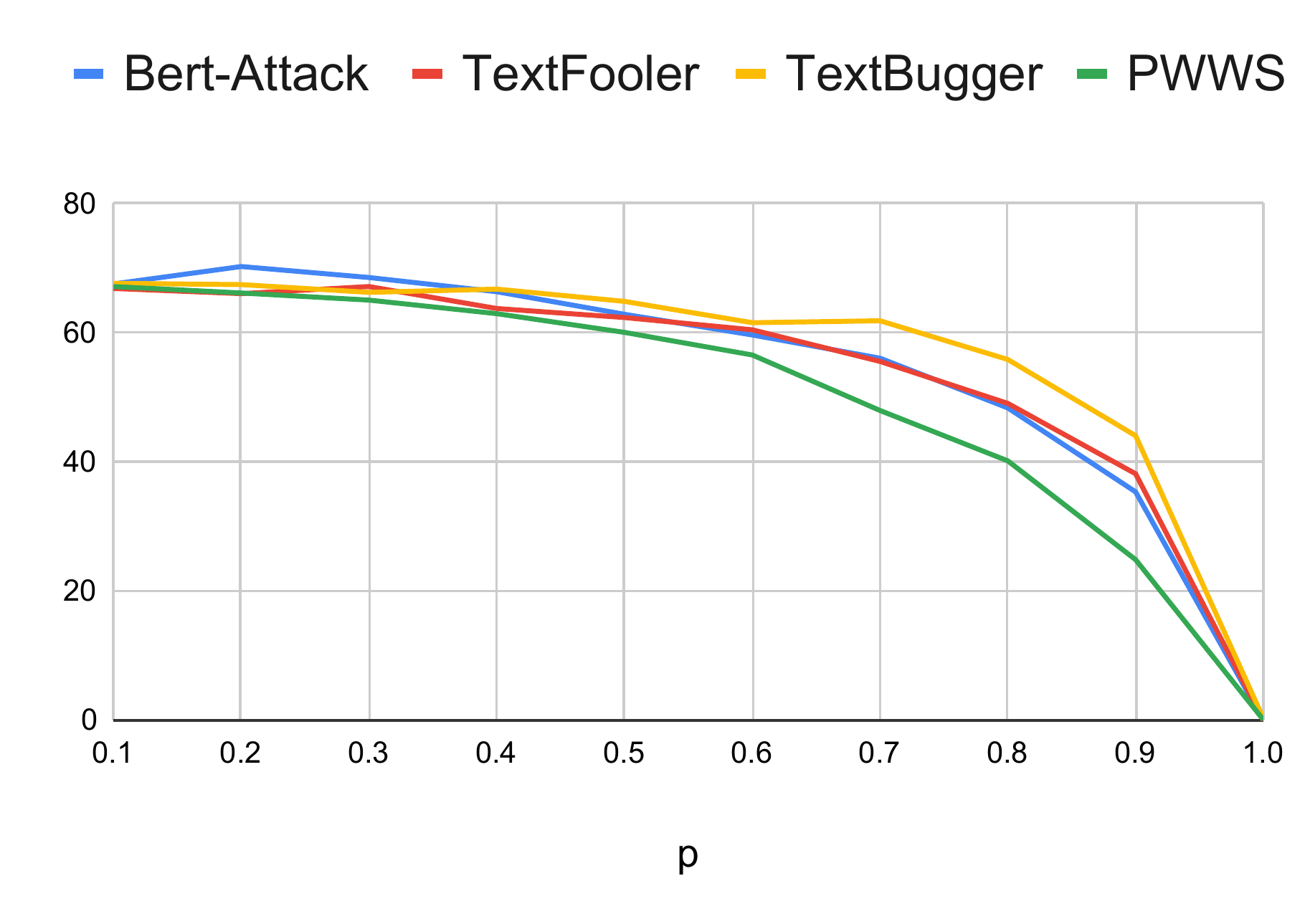}
    \caption{Accuracy with various p values for LSTM on IMDB. Note that $k$ is fixed to 100.}
    \label{fig:optimalp}
\end{figure}

\begin{figure}
    \centering
    \includegraphics[width=0.8\columnwidth]{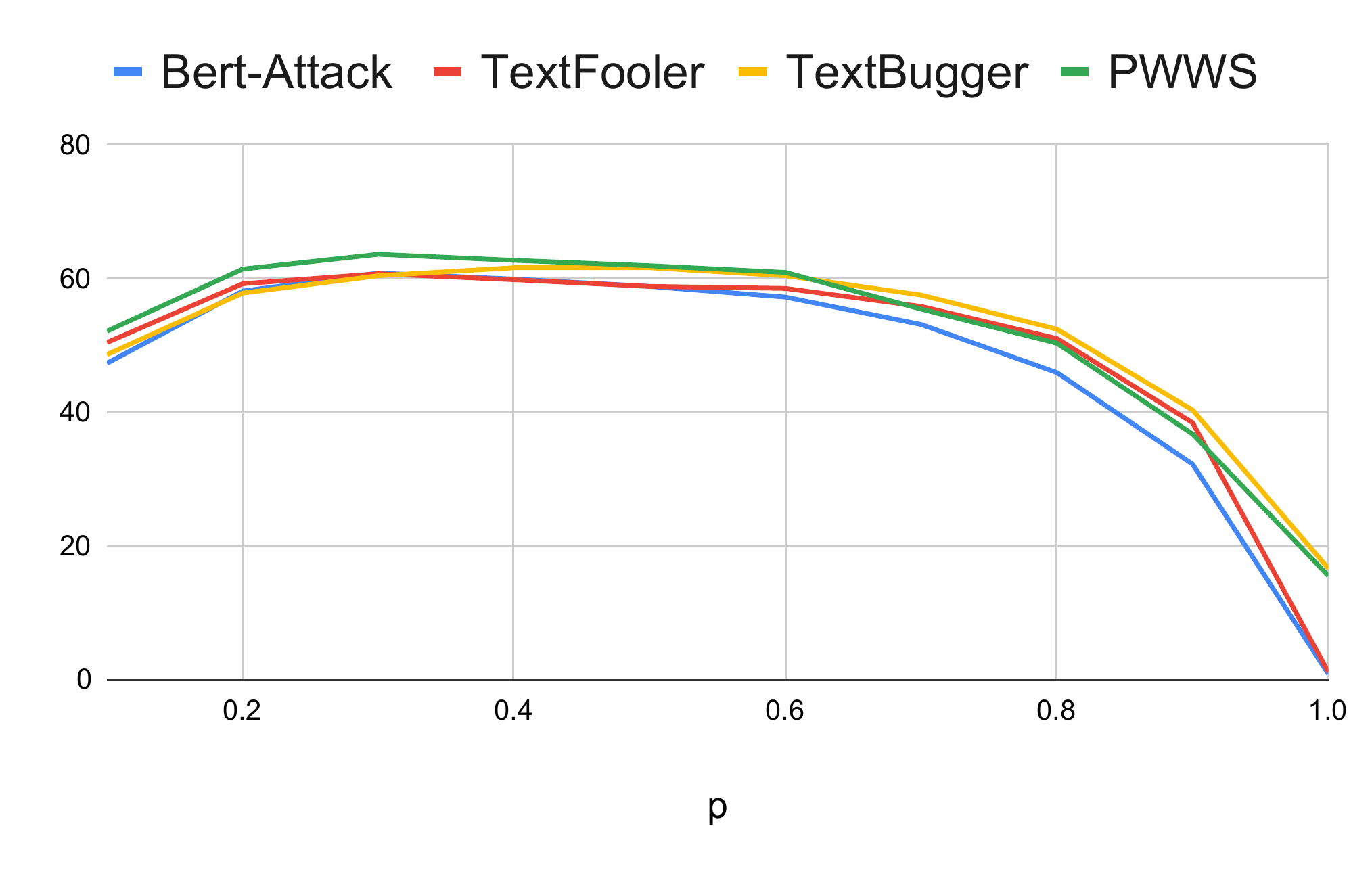}
    \caption{Accuracy with various p values for LSTM on AG News. Note that $k$ is fixed to 100.}
    \label{fig:agoptimalp}
\end{figure}

\noindent \textbf{Optimal $p$.} Figure \ref{fig:optimalp} shows the results for all combinations of attacks against LSTM on IMDB with word shielding as the defense, $k$ fixed at 100. As we increase $p$, we see a continued drop in accuracy which is consistent with the idea that a higher $p$ is more likely to capture perturbed text. The optimal value range appears to be in 0.2 - 0.4 range, although we do not see large drops until 0.6 onward. We also examined the same combination on AG News (Figure \ref{fig:agoptimalp}) since it's texts are considerably shorter and found consistent results.

\begin{figure}
    \centering
    \includegraphics[width=0.8\columnwidth]{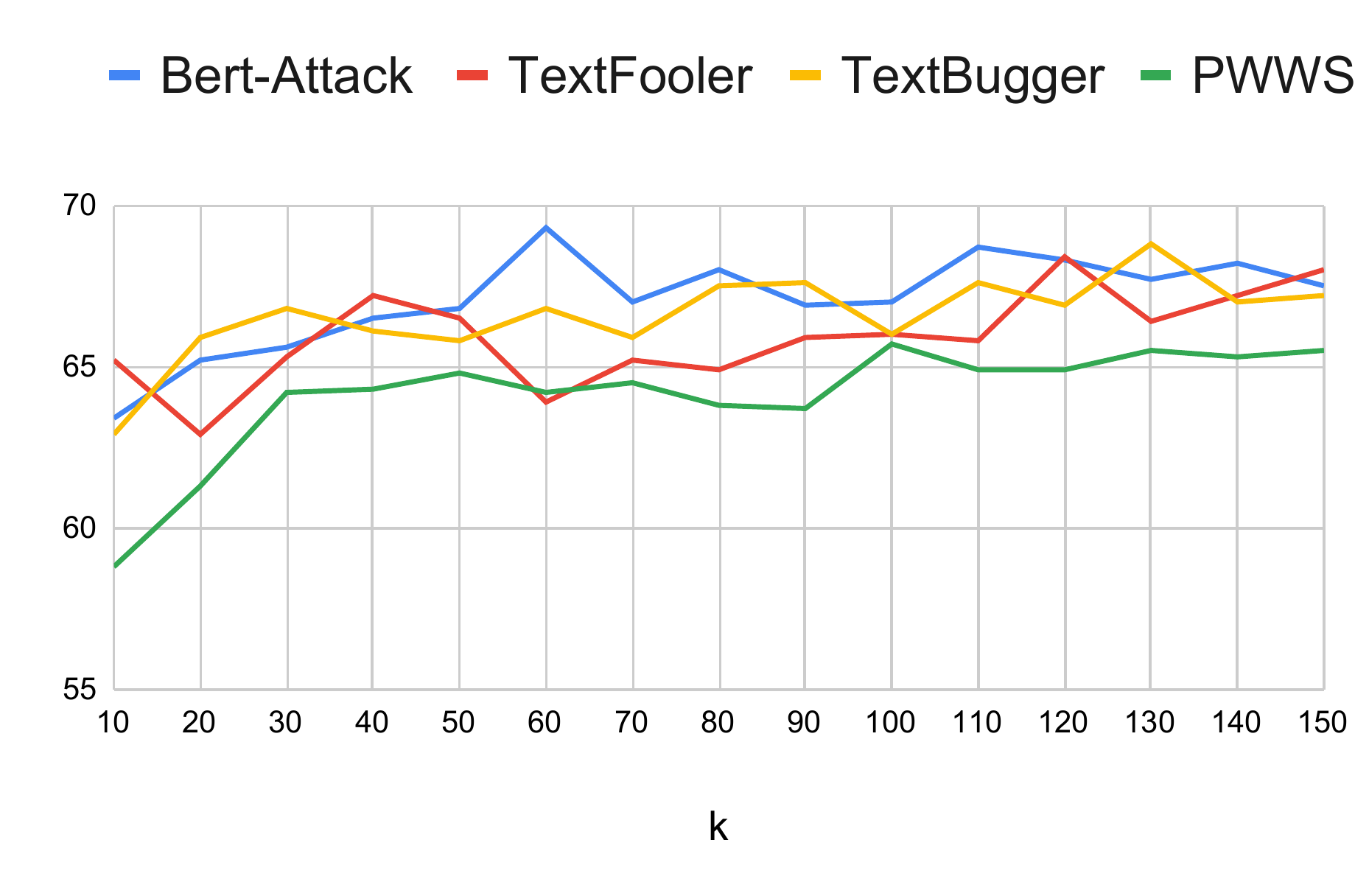}
    \caption{Accuracy with various k values for LSTM on IMDB. Note that $p$ is fixed to 0.3.}
    \label{fig:optimalk}
\end{figure}

\begin{figure}
    \centering
    \includegraphics[width=0.8\columnwidth]{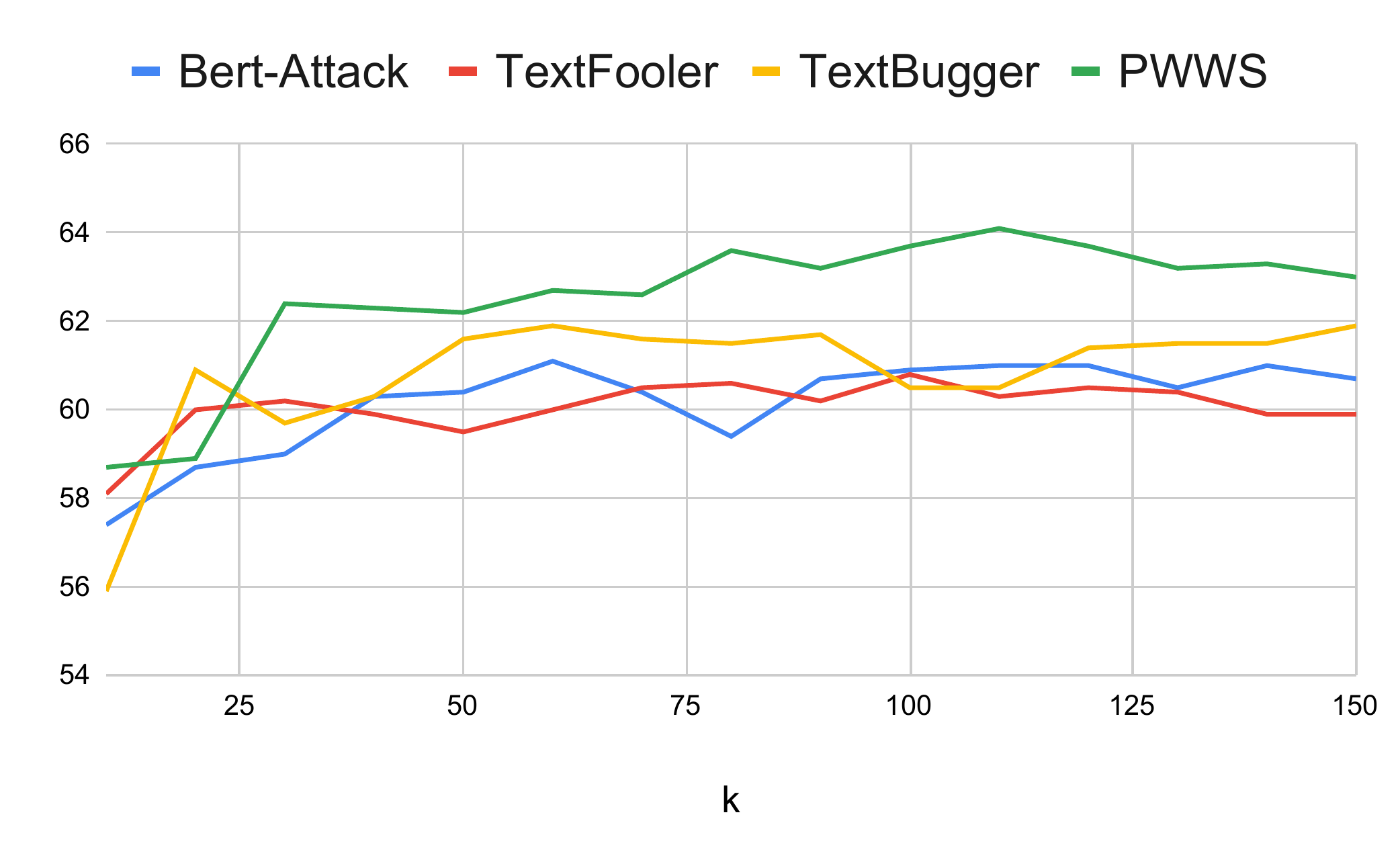}
    \caption{Accuracy with various k values for LSTM on AG News. Note that $p$ is fixed to 0.3.}
    \label{fig:agoptimalk}
\end{figure}

\noindent \textbf{Optimal $k$.} Figure \ref{fig:optimalk} shows results for all attacks against LSTM on IMDB with word sampling as the defense, $p$ fixed at 0.3. The optimal $k$ is not as clear as $p$. We see clear increases after 30 samples, but then the optimal $k$ varies depending on attack. However, we see a leveling off around 90 samples, which gives some credence to our chosen $k$ of 100. We also found similar results when examining the same combination on AG News (Figure \ref{fig:agoptimalk}), however, $k$ stabilized lower (about 50).

\subsection{Reliability of \textit{Sample Shielding}}

\begin{figure}
    \centering
    \includegraphics[width=0.7\columnwidth]{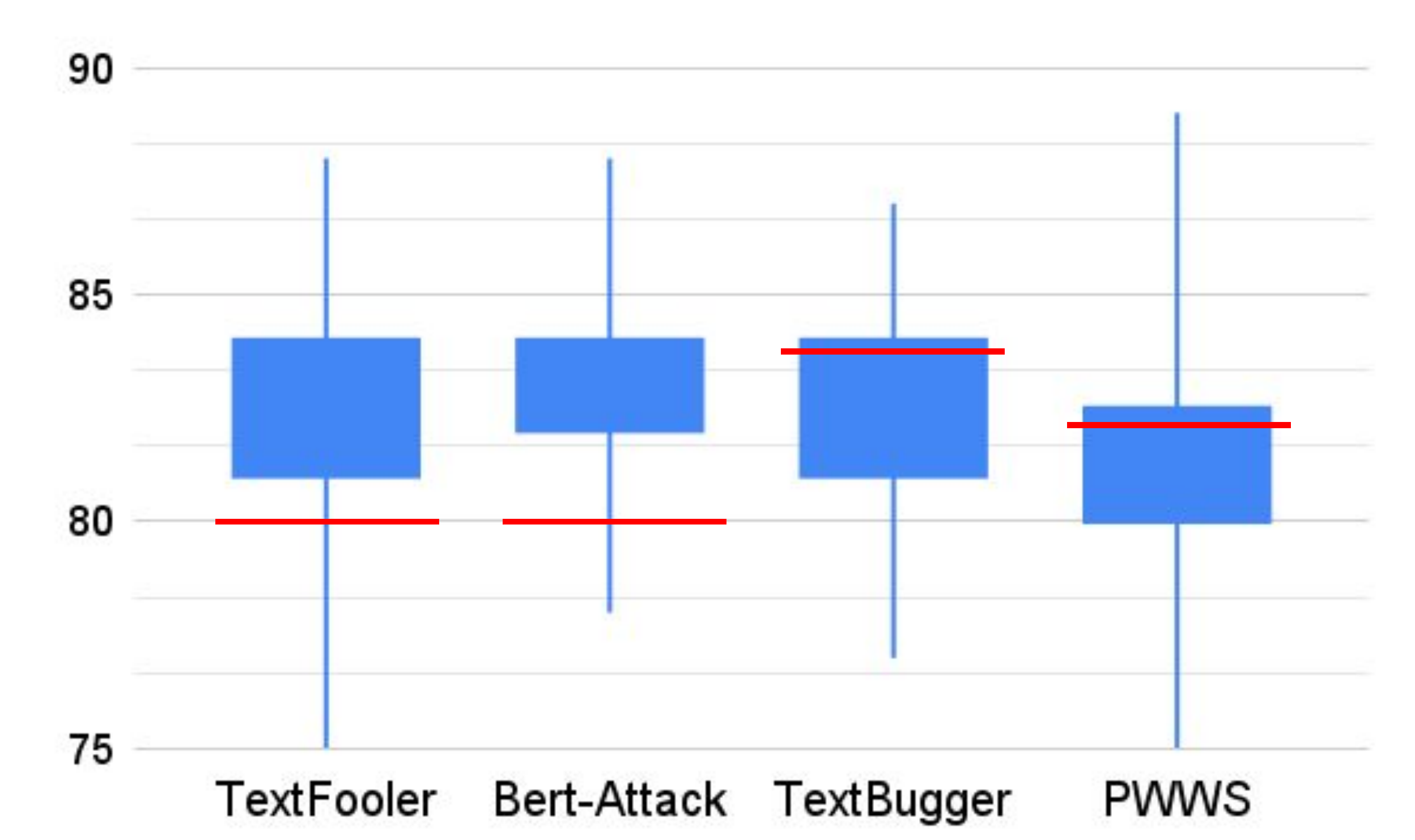}
    \caption{Boxplots of accuracies when \textit{Sample Shielding} is applied 100 times to attacked IMDB texts with BERT as classifier. Red lines: accuracies reported in Table \ref{tab:adversarialResults}. }
    \label{fig:NApplications}
\end{figure}

Due to the randomness of samples, there may be concern over the consistency of \emph{Sample Shielding}. To address this, we ran \emph{Sample Shielding} 100 times on the IMDB attacked texts from Table \ref{tab:adversarialResults} against BERT classifier.
Each time 30 random samples were used to vote. As can be observed from Figure \ref{fig:NApplications},  \textit{Sample Shielding} consistently protects against attacks. Median accuracies are above 80\% dropping only to 75\% in the worst case. This points to Sample Shielding as a consistent, reliable defense.

\begin{table*}[]
    \centering
    \footnotesize
    \begin{tabular}{c||c||c|c||c|c||c|c||c|c||c|c}
        & & Sample & Orig. & \multicolumn{2}{|c|}{TextFooler} & \multicolumn{2}{|c|}{Bert-Attack} & \multicolumn{2}{|c|}{TextBugger} & \multicolumn{2}{|c}{PWWS} \\
         & Classifier & Strategy & Acc. & Acc. & ASR  & Acc. & ASR & Acc. & ASR & Acc.  & ASR \\\hline
         \parbox[t]{2mm}{\multirow{5}{*}{\rotatebox[origin=c]{90}{IMDB}}}  &  LSTM & Local (W') Word & 87 &	11 & 87	& 31 & 64	& 28 & 68 & 22 & 75\\\cline{2-12}
         
         & CNN & Local (W') Word & 91 & 22 & 76	& 22 & 76	& 32 & 65 & 29 & 68\\\cline{2-12}
         
         & BERT & Local (W') Word & 81 & 18 & 78	& 31 & 62	& 34 & 58 & 26 & 68\\\cline{2-12} \cline{2-12} 
         
         & BERT & RanMask* & 92 & 22 & 75 & 36 & 58 & 18 & 79  & - & - \\
         
         & BERT & FreeLB++* & 93 & 45 & 51 & 40 & 57 & 43 & 54 & - & -\\\hline

         
         

          \parbox[t]{2mm}{\multirow{5}{*}{\rotatebox[origin=c]{90}{AG }}} 
         & LSTM & Local (W') Word & 88 & 42 & 52 & 31 & 65	& 38 & 57 & 55 & 38\\\cline{2-12}
         
         & CNN &  Local (W') Word & 86 & 45 & 48 & 28 & 67	& 36 & 58 & 54 & 37 \\\cline{2-12} 
         
         & BERT & Local (W') Word & 88 & 48 & 45	& 38 & 57	& 55 & 38 & 64 & 27\\\cline{2-12}\cline{2-12}
         
        & BERT & RanMask* & 92 & 38 &  59 & 49 & 46 & 45 & 51 & - & - \\
         
        & BERT & FreeLB++* & 95 & 52 & 46  & 42 & 56 & 56 & 41 & - & -\\\hline
         
    \end{tabular}
    \caption{Results of attack against local model with knowledge of \textit{Sample Shielding}. For all shielding cases, $k=30$, $p=0.3$, and majority voting is used. Acc. is accuracy, and ASR is success rate of attack (\%) and Orig. Acc. is accuracy on the original text. Note that the examples used by RanMask and FreeLB++ is not the set of dataset samples as our paper. }
    \label{tab:localAdversarialResults}
\end{table*}

\subsection{Comparison with other SOTA Defenses}

Comparisons are limited as threat models differ. 
As noted earlier, other defenses assume a weaker threat model where the attacker queries the website's shielded $W$ directly. 
To make ours equivalent we compare SOTA results with our accuracies obtained by the attacker using $W'$ alone (with $W = W'$).
We calculate accuracies right after the final perturbed text is generated using $W'$ eliminating a followup round of $W$ with \textit{Sample Shielding}. Table \ref{tab:localAdversarialResults} provides our full results against this weaker threat model.

With BERT as base classifier for AG News, FreeLB++, an adversarial training technique \cite{Li2021SearchingFA} report accuracies of 51, 56, and 42 against TextFooler, TextBugger, and Bert-Attack respectively.  RanMask \cite{Zeng2021CertifiedRT}, which uses random masking of words report accuracies of 38, 45, and 49. 
In comparison, \textit{Sample Shielding} achieves 48, 55, and 38 respectively outperforming RanMask in 2 out of 3, while only a fews point behind FreeLB++. 
For IMDB, FreeLB++ reports 45, 43, and 40 and RanMask reports 22, 18, and 36 respectively.  Equivalently, \textit{Sample Shielding} achieves 18, 34, and 31.
With some wins and some losses, \textit{Sample Shielding} is in the mix with current SOTA defenses in this weaker threat model.
\emph{However, when deployed as designed for the realistic threat model, it wins over these other defenses by large margins (see Table \ref{tab:adversarialResults})}.
While we do not know how FreeLB++,  RanMask, and similar defenses would perform with our threat model any deterministic shield would give the exact same results when the classifier is applied once again by the website.

\subsection{Limitations/Future work}

First, in future work we will add in direct comparisons to the two closest methods to \textit{Sample Shielding} \cite{Zeng2021CertifiedRT, Wang2021RandomizedSA}. They are similar in spirit as they also work off samples though these are generated differently. We have not compared with them because these two papers appeared very recently, one last revised in July \cite{Zeng2021CertifiedRT} and the other appeared in arXiv in September 2021 \cite{Wang2021RandomizedSA}. Second, the neural net summarizer leverages a simple linear layer. Other networks, e.g.,  LSTM, maybe better at finding patterns in sequential data. In future work we will also explore layering \textit{Sample Shielding} onto other defense strategies. 

Another limitation of our current method is that we do not measure Sample Shielding's effectiveness on other common text tasks including Natural Language Understanding. Additionally, datasets which contain the shortest texts (e.g. SST2) are not currently tested in our experiments. Since sample shielding removes texts, it's performance could drop for these tasks and short texts. Thus, future work will include these comparisons.

\section{Related Work} 


\noindent \textbf{Defenses using voting.} The most similar methods to our own are RanMask and RS\&V both appearing within the last five months. RanMask \cite{Zeng2021CertifiedRT} randomly masks tokens in input texts. This random masking occurs $n$ times generating $n$ inputs to be fed to a classifier.
RS\&V \cite{Wang2021RandomizedSA} randomly replaces words in the input with synonyms. This it does $k$ times to produce $k$ samples which are then voted on. If the samples vote for a different label than the label produced by the unsampled input, then the text is labeled as an adversarial text. 
Our method is advantageous since it does not rely on specific models (i.e. Masked Language Model) or synonym sources. 


\noindent \textbf{Adversarial training.} Classifiers train on perturbed data, learning to identify modified versions of the original input \cite{Wang2020DefenseOW, Wang2021AdversarialTW, Zhu2020FreeLB, Li2021SearchingFA}. As an example, 
\citet{gil2019whitetoblack} propose HotFlip which uses white-box knowledge to generate adversarial attacks to train on. Specifically, they flip tokens based on the gradients of the one-hot input vectors.
%
%
%
%
%
%
However, adversarial defenses are limited to known attackers.
%
%
In contrast, \textit{Sample Shielding} is `plug-and-play' as it is a pre-processing step.

\noindent \textbf{Other defenses.} Several other shielding methods exist \cite{keller-etal-2021-bert, eger-etal-2019-text, Zhu2021TREATEDTU}.
For example, \citet{rodriguez2018google} defend Perspective (Google's toxicity classification model) by neutralizing adversarial inputs via a negated predicates list. 
%
%
%
%
Again, these defenses are restricted to contexts where specific lists may be identified, this is not so with \textit{Sample Shielding}.

\section{Conclusion}


\textit{Sample Shielding}, an intuitively designed defense which is attacker and classifier agnostic, protects effectively; reducing ASR from 90 - 100\% down to 14 - 34\% with minimal accuracy loss (3\%) in original texts. The randomness (through sampling) provides unreliable feedback for attackers, thus it even thwarts attackers who have query access to   classifiers protected with \textit{Sample Shielding}. 
Attack strategies will need to increase the amount of perturbation to make sure a majority of samples fail at classification. However, this will risk semantic integrity. Thus, we expect \textit{Sample Shielding} to cause ripples in future adversarial attack strategies while providing text classifiers with a definite advantage.

\bibliography{main}

\begin{thebibliography}{22}
\expandafter\ifx\csname natexlab\endcsname\relax\def\natexlab#1{#1}\fi

\bibitem[{Eger et~al.(2019)Eger, {\c{S}}ahin, R{\"u}ckl{\'e}, Lee, Schulz,
  Mesgar, Swarnkar, Simpson, and Gurevych}]{eger-etal-2019-text}
Steffen Eger, G{\"o}zde~G{\"u}l {\c{S}}ahin, Andreas R{\"u}ckl{\'e}, Ji-Ung
  Lee, Claudia Schulz, Mohsen Mesgar, Krishnkant Swarnkar, Edwin Simpson, and
  Iryna Gurevych. 2019.
\newblock \href {https://doi.org/10.18653/v1/N19-1165} {Text processing like
  humans do: Visually attacking and shielding {NLP} systems}.
\newblock In \emph{Proceedings of the 2019 Conference of the North {A}merican
  Chapter of the Association for Computational Linguistics: Human Language
  Technologies, Volume 1 (Long and Short Papers)}, pages 1634--1647,
  Minneapolis, Minnesota. Association for Computational Linguistics.

\bibitem[{Garg and Ramakrishnan(2020)}]{garg2020bae}
Siddhant Garg and Goutham Ramakrishnan. 2020.
\newblock \href {http://arxiv.org/abs/2004.01970} {Bae: Bert-based adversarial
  examples for text classification}.

\bibitem[{Gil et~al.(2019)Gil, Chai, Gorodissky, and
  Berant}]{gil2019whitetoblack}
Yotam Gil, Yoav Chai, Or~Gorodissky, and Jonathan Berant. 2019.
\newblock \href {https://doi.org/10.18653/v1/N19-1139} {White-to-black:
  Efficient distillation of black-box adversarial attacks}.
\newblock In \emph{Proceedings of the 2019 Conference of the North {A}merican
  Chapter of the Association for Computational Linguistics: Human Language
  Technologies, Volume 1 (Long and Short Papers)}, pages 1373--1379,
  Minneapolis, Minnesota. Association for Computational Linguistics.

\bibitem[{Goel et~al.(2020)Goel, Agarwal, Vatsa, Singh, and
  Ratha}]{goal2020DNDNet}
Akhil Goel, Akshay Agarwal, Mayank Vatsa, Richa Singh, and Nalini~K. Ratha.
  2020.
\newblock \href {https://doi.org/10.1109/CVPRW50498.2020.00019} {Dndnet:
  Reconfiguring cnn for adversarial robustness}.
\newblock In \emph{2020 IEEE/CVF Conference on Computer Vision and Pattern
  Recognition Workshops (CVPRW)}, pages 103--110.

\bibitem[{Jia et~al.(2019)Jia, Raghunathan, Göksel, and
  Liang}]{jia2019certified}
Robin Jia, Aditi Raghunathan, Kerem Göksel, and Percy Liang. 2019.
\newblock \href {http://arxiv.org/abs/1909.00986} {Certified robustness to
  adversarial word substitutions}.

\bibitem[{Jin et~al.(2020)Jin, Jin, Zhou, and Szolovits}]{jin2020bert}
Di~Jin, Zhijing Jin, Joey~Tianyi Zhou, and Peter Szolovits. 2020.
\newblock Is bert really robust? a strong baseline for natural language attack
  on text classification and entailment.
\newblock In \emph{Proceedings of the AAAI conference on artificial
  intelligence}, volume~34, pages 8018--8025.

\bibitem[{Keller et~al.(2021)Keller, Mackensen, and
  Eger}]{keller-etal-2021-bert}
Yannik Keller, Jan Mackensen, and Steffen Eger. 2021.
\newblock \href {https://doi.org/10.18653/v1/2021.findings-acl.141}
  {{BERT}-defense: A probabilistic model based on {BERT} to combat cognitively
  inspired orthographic adversarial attacks}.
\newblock In \emph{Findings of the Association for Computational Linguistics:
  ACL-IJCNLP 2021}, pages 1616--1629, Online. Association for Computational
  Linguistics.

\bibitem[{Kim(2014)}]{kim-2014-convolutional}
Yoon Kim. 2014.
\newblock \href {https://doi.org/10.3115/v1/D14-1181} {Convolutional neural
  networks for sentence classification}.
\newblock In \emph{Proceedings of the 2014 Conference on Empirical Methods in
  Natural Language Processing ({EMNLP})}, pages 1746--1751, Doha, Qatar.
  Association for Computational Linguistics.

\bibitem[{Li et~al.(2021{\natexlab{a}})Li, Zhang, Peng, Chen, Brockett, Sun,
  and Dolan}]{li2021contextualized}
Dianqi Li, Yizhe Zhang, Hao Peng, Liqun Chen, Chris Brockett, Ming-Ting Sun,
  and Bill Dolan. 2021{\natexlab{a}}.
\newblock \href {https://doi.org/10.18653/v1/2021.naacl-main.400}
  {Contextualized perturbation for textual adversarial attack}.
\newblock In \emph{Proceedings of the 2021 Conference of the North American
  Chapter of the Association for Computational Linguistics: Human Language
  Technologies}, pages 5053--5069, Online. Association for Computational
  Linguistics.

\bibitem[{Li et~al.(2019)Li, Ji, Du, Li, and Wang}]{Li2019Textbugger}
Jinfeng Li, Shouling Ji, Tianyu Du, Bo~Li, and Ting Wang. 2019.
\newblock \href {https://doi.org/10.14722/ndss.2019.23138} {Textbugger:
  Generating adversarial text against real-world applications}.
\newblock \emph{Proceedings 2019 Network and Distributed System Security
  Symposium}.

\bibitem[{Li et~al.(2020)Li, Ma, Guo, Xue, and Qiu}]{li-etal-2020-bert-attack}
Linyang Li, Ruotian Ma, Qipeng Guo, Xiangyang Xue, and Xipeng Qiu. 2020.
\newblock \href {https://doi.org/10.18653/v1/2020.emnlp-main.500}
  {{BERT}-{ATTACK}: Adversarial attack against {BERT} using {BERT}}.
\newblock In \emph{Proceedings of the 2020 Conference on Empirical Methods in
  Natural Language Processing (EMNLP)}, pages 6193--6202, Online. Association
  for Computational Linguistics.

\bibitem[{Li et~al.(2021{\natexlab{b}})Li, Xu, Zeng, Li, Zheng, Zhang, Chang,
  and Hsieh}]{Li2021SearchingFA}
Zongyi Li, Jianhan Xu, Jiehang Zeng, Linyang Li, Xiaoqing Zheng, Qi~Zhang,
  Kai-Wei Chang, and Cho-Jui Hsieh. 2021{\natexlab{b}}.
\newblock Searching for an effiective defender: Benchmarking defense against
  adversarial word substitution.
\newblock In \emph{EMNLP}.

\bibitem[{Morris et~al.(2020)Morris, Lifland, Yoo, Grigsby, Jin, and
  Qi}]{morris2020textattack}
John Morris, Eli Lifland, Jin~Yong Yoo, Jake Grigsby, Di~Jin, and Yanjun Qi.
  2020.
\newblock \href {https://doi.org/10.18653/v1/2020.emnlp-demos.16}
  {{T}ext{A}ttack: A framework for adversarial attacks, data augmentation, and
  adversarial training in {NLP}}.
\newblock In \emph{Proceedings of the 2020 Conference on Empirical Methods in
  Natural Language Processing: System Demonstrations}, pages 119--126, Online.
  Association for Computational Linguistics.

\bibitem[{Ren et~al.(2019)Ren, Deng, He, and Che}]{ren-etal-2019-generating}
Shuhuai Ren, Yihe Deng, Kun He, and Wanxiang Che. 2019.
\newblock \href {https://doi.org/10.18653/v1/P19-1103} {Generating natural
  language adversarial examples through probability weighted word saliency}.
\newblock In \emph{Proceedings of the 57th Annual Meeting of the Association
  for Computational Linguistics}, pages 1085--1097, Florence, Italy.
  Association for Computational Linguistics.

\bibitem[{Rodriguez and Galeano(2018)}]{rodriguez2018google}
Nestor Rodriguez and Sergio~Rojas Galeano. 2018.
\newblock \href {http://arxiv.org/abs/1801.01828} {Shielding google's language
  toxicity model against adversarial attacks}.
\newblock \emph{CoRR}, abs/1801.01828.

\bibitem[{Wang et~al.(2021{\natexlab{a}})Wang, Xiong, and
  He}]{Wang2021RandomizedSA}
Xiaosen Wang, Yifeng Xiong, and Kun He. 2021{\natexlab{a}}.
\newblock Randomized substitution and vote for textual adversarial example
  detection.
\newblock \emph{ArXiv}, abs/2109.05698.

\bibitem[{Wang et~al.(2021{\natexlab{b}})Wang, Yang, Deng, and
  He}]{Wang2021AdversarialTW}
Xiaosen Wang, Yichen Yang, Yihe Deng, and Kun He. 2021{\natexlab{b}}.
\newblock Adversarial training with fast gradient projection method against
  synonym substitution based text attacks.
\newblock In \emph{AAAI}.

\bibitem[{Wang and Wang(2020)}]{Wang2020DefenseOW}
Zhaoyang Wang and Hongtao Wang. 2020.
\newblock Defense of word-level adversarial attacks via random substitution
  encoding.
\newblock In \emph{KSEM}.

\bibitem[{Yoo and Qi(2021)}]{yoo2021improving}
Jin~Yong Yoo and Yanjun Qi. 2021.
\newblock \href {http://arxiv.org/abs/2109.00544} {Towards improving
  adversarial training of nlp models}.

\bibitem[{Zeng et~al.(2021)Zeng, Zheng, Xu, Li, Yuan, and
  Huang}]{Zeng2021CertifiedRT}
Jiehang Zeng, Xiaoqing Zheng, Jianhan Xu, Linyang Li, Liping Yuan, and Xuanjing
  Huang. 2021.
\newblock Certified robustness to text adversarial attacks by randomized
  [mask].
\newblock \emph{ArXiv}, abs/2105.03743.

\bibitem[{Zhu et~al.(2021)Zhu, Gu, Wang, and Tian}]{Zhu2021TREATEDTU}
Bin Zhu, Zhaoquan Gu, Le~Wang, and Zhihong Tian. 2021.
\newblock Treated: Towards universal defense against textual adversarial
  attacks.
\newblock \emph{ArXiv}, abs/2109.06176.

\bibitem[{Zhu et~al.(2020)Zhu, Cheng, Gan, Sun, Goldstein, and
  Liu}]{Zhu2020FreeLB}
Chen Zhu, Yu~Cheng, Zhe Gan, Siqi Sun, Tom Goldstein, and Jingjing Liu. 2020.
\newblock \href {https://openreview.net/forum?id=BygzbyHFvB} {Freelb: Enhanced
  adversarial training for natural language understanding}.
\newblock In \emph{International Conference on Learning Representations}.

\end{thebibliography}
\bibliographystyle{acl_natbib}

\end{document}